\documentclass[10pt, a4paper]{article}
\pdfoutput=1
\usepackage{colortbl}

\usepackage{emnlp2023}

\usepackage{times}
\usepackage{latexsym}
\usepackage{graphicx}
\usepackage{booktabs}

\usepackage[T2A,T1]{fontenc}

\usepackage[utf8]{inputenc}
\usepackage[english,ukrainian]{babel}
\usepackage[babel=true]{microtype}
\usepackage{tempora}

\usepackage{caption}
\usepackage{inconsolata}
\usepackage{amsmath}
\usepackage{amsfonts}
\usepackage{bbm}
\usepackage{listings}
\usepackage{float}
\usepackage{todonotes}
\usepackage{gb4e} 
\usepackage{multirow} 
\usepackage{enumitem}

\newcommand{\ukr}[1]{{\it #1}}
\newcommand{\csuk}{CS$\rightarrow$UK}
\newcommand{\ukcs}{UK$\rightarrow$CS}

\newcommand{\CT}{Charles Translator}
\newcommand{\lindaturl}{\url{https://lindat.mff.cuni.cz/services/translation/}}

\definecolor{ashgrey}{rgb}{0.7, 0.75, 0.71}

\let\origfootnote\footnote
\renewcommand{\footnote}[1]{\kern.1em\origfootnote{#1}}
\newcommand{\punctfootnote}[1]{\kern-.1em\origfootnote{#1}}

\title{Charles Translator: A Machine Translation System between Ukrainian and Czech}

\author{
Martin Popel$^1$ \quad
Lucie Poláková$^1$ \quad
Michal Novák$^1$ \quad
Jindřich Helcl$^1$ \\
\large\bfseries
Jindřich Libovický$^1$ \quad
Pavel Straňák$^1$ \quad
Tomáš Krabač$^{1}$ \\
\large\bfseries
Jaroslava Hlaváčová$^1$ \quad
Mariia Anisimova$^1$ \quad
Tereza Chlaňová$^2$ \\
$^1$Charles University, Faculty of Mathematics and Physics, Institute of Formal and Applied Linguistics \\
$^2$Charles University, Faculty of Arts, Department of East European Studies\\
Prague, Czech Republic \\
{\normalsize \fontencoding{T1}\selectfont \tt
\{surname\}@ufal.mff.cuni.cz, mnovak@ufal.mff.cuni.cz, tereza.chlanova@ff.cuni.cz}}

\begin{document}
\maketitle

\begin{abstract}
We present \CT{}, a machine translation system between Ukrainian and Czech, developed as part of a society-wide effort to mitigate the impact of the
Russian-Ukrainian war
on individuals and society.
The system was developed in the spring of 2022 with the help of many language data providers in order to quickly meet the demand for such a service, which was not available at the time in the required quality.
The translator was later implemented as an online web interface and as an Android app with speech input, both featuring Cyrillic-Latin script transliteration. The system translates directly, compared to other available systems that use English as a pivot, and thus take advantage of the typological similarity of the two languages.
It uses the block back-translation method, which allows for efficient use of monolingual training data.
The paper describes the development process, including data collection and implementation, evaluation, mentions several use cases, and outlines possibilities for the further development of the system for educational purposes.
\end{abstract}

\section{Introduction}
\label{sec:intro}

As a result of the Russian invasion of Ukraine in February 2022, the Czech Republic became one of the main countries to host people forced to flee their homes. According to sources from the UN High Commissioner for Refugees (UNHCR), it is the fourth country with the largest number of Ukrainian refugees. By April 1, 2023, more than 504,000 Ukrainians had been granted temporary protection in the country, of whom more than 325,000 had applied for an extension of their refugee status beyond March 2023.%
\punctfootnote{%
\url{https://data2.unhcr.org/en/documents/details/104052}
}
Virtually overnight, there arose the need for a fast and effective means of communication between Czech and Ukrainian speakers, which until then did not have the required quality.

Our motivation to develop such a service, apart from the wish to help reduce the language (and social) barrier between the refugees and the Czech society, is based on several convenient factors: (i)~our previous long-term scientific experience in the field of machine translation (MT) and the existence of an appropriate MT method,
(ii)~the proximity of the two Slavic languages in question, and (iii)~the availability of resources: the possibility of obtaining training data from multiple volunteer subjects and the willingness of many researchers to prioritize this line of research, leading to a quick solution with a quick implementation process.

The translation systems available to the public during the conflict outbreak translated only indirectly between Czech and Ukrainian by pivoting through English. This approach does not take advantage of the typological affinity of the two languages, such as the high inflection with rich morphology enabling great flexibility of word order, pro-drop, partial lexical similarity, e.g. \textit{můj dům} -- \ukr{мій дім} (my house), \textit{chladná zima} -- \ukr{xолодна зима} (cold winter), \textit{krátké vlasy} -- \ukr{коротке волосся} (short hair)\footnote{At the same time, there is quite a large number of false friends, e.g.: \ukr{квітень} -- \textit{duben} (April, resembling \textit{květen} -- \textit{May} in Czech); \ukr{лікарня} -- nemocnice (hospital, resembling \textit{lékárna} -- \textit{pharmacy} in Czech), \ukr{напад} -- \textit{útok} (attack, resembling \textit{nápad} -- \textit{idea} in Czech).} and syntactic similarities.




With English as the pivot, some information is inevitably lost. This is most visible in the grammatical categories of gender and politeness: both Ukrainian and Czech distinguish masculine, feminine, and neutral forms of nouns and adjectives, and a formal (\textit{Vy}, \ukr{Ви}) and an informal (\textit{ty}, \ukr{ти}) form of the \textit{you}-pronoun (in singular and plural). Thus, the following Czech example 
with a female speaker addressing the hearer in a formal (polite) way:

\vspace{-5mm}
\begin{quote}
  \gll Jsem~nemocná.~A~co~Vy?\\
  0-am~sick-\textbf{F}.SG.~And~what~you-\textbf{FORMAL}.SG?\\
  \trans `I'm sick. And you?'
\end{quote}
%
gets incorrectly translated as a male speaker informally addressing the listener:

\begin{quote}
  \gll \ukr{Я хворий. А ти?}\\
  Ja~chvoryj.~A~ty?\\
  I sick-\textbf{M}.SG. And you-\textbf{INFORMAL}.SG?\\
\end{quote}
\vspace{-5mm}
%
Similarly, there also can be a fatal shift of meaning when, for instance, the Czech \textit{Jaké léky to jsou?} (\textit{What medicine is that?}) translates as \ukr{Що це за наркотики?} (literally \textit{What are these narcotics?}) since English \textit{drugs} can mean both \textit{medicine} and \textit{drugs}/\textit{narcotics}.
\CT{} translates directly, and thus is not prone to these errors.


\def\SEC{§}
In the rest of the paper, we present related work (\SEC~\ref{sec:relwork});
 the translator architecture, training and test data, deployment and user interfaces (\SEC~\ref{sec:components});
 evaluation (\SEC~\ref{sec:quality});
 use cases and usage statistics (\SEC~\ref{sec:usecases})
 and we conclude with plans for future (\SEC~\ref{sec:conclusions}).



%

\section{Related Work}
\label{sec:relwork}

Current MT methods are largely language independent and rely on the Transformers architecture \citep{vaswani2017attention}, making substantial use of back-translation \citep{sennrich-etal-2016-improving,edunov-etal-2018-understanding} and data filtering in all stages of model training \citep{junczys-dowmunt-2018-dual}.

Translation between Czech and Ukrainian was part of the WMT22 evaluation campaign \citep{kocmi-etal-2022-findings}, with most participants relying on language-agnostic methods. The winning system \citep{nowakowski-etal-2022-adam} enriched the source side of the translation with information about named entities and used complex decoding with a neural model to restore hypotheses. In our submission \citet{popel-etal-2022-cuni}, we used handcrafted regular expressions to handle errors in named entities during data filtering, which is also part of \CT{}. Similarly to \citet{alabi-etal-2022-inria}, we experimented with romanization, which is not used in the deployed system.

Although the implementation aspects of MT are discussed in the literature \citep{junczys-dowmunt-etal-2018-marian,behnke-etal-2021-efficient,heafield-etal-2022-findings}, there is virtually no related work focusing on the deployment of machine translation, which typically remains part of the secret know-how of commercial MT providers.

\section{Components of the Translator}
\label{sec:components}

\CT{} consists of the translation service and multiple interfaces for accessing the translator.

\subsection{The translation service}
\label{method-etc}
\label{sec:testsets} 

\paragraph{Method.}
We use the Transformer architecture \citep{vaswani2017attention} with iterated block back-translation \citep{popel-et-al:2020}, allowing for more efficient monolingual training data use. The system was trained in the same way as the sentence-level English-Czech system of \citet{popel-2018-cuni}.

\paragraph{Training Data.}

The collection of training data for the first model took place over a short and intensive period with the help of a wide range of volunteer subjects. Cooperation with Czech-Ukrainian translators, translation agencies, and the authors of the InterCorp parallel corpus \citeplanguageresource{InterCorp}, a project of the Czech National Corpus, was important for obtaining good quality parallel data.

We also used data available at the OPUS repository \citeplanguageresource{tiedemann-2012-parallel}, namely texts from the Bible \citeplanguageresource{christodouloupoulos2015massively},
CCMatrix \citeplanguageresource{schwenk-etal-2019-ccmatrix,fan-etal-2021-beyond}, ELRC, EUBookshop, GNOME, KDE4,
MultiCCAligned \citeplanguageresource{elkishky-ccaligned-2020},
MultiParaCrawl,\punctfootnote{\url{https://paracrawl.eu}}
OpenSubtitles \citeplanguageresource{lison-tiedemann-2016-opensubtitles2016},
QED \citep{abdelali-etal-2014-amara},
Tatoeba, TED2020 \citeplanguageresource{reimers-gurevych-2020-making},
Ubuntu, WikiMatrix \citeplanguageresource{schwenk-etal-2021-wikimatrix}, and XLEnt \citeplanguageresource{el-kishky-etal-2021-xlent}.

In addition to the (authentic) parallel data,
 we also used monolingual data for backtranslation:
 50M originally Czech sentences from CzEng~2.0 \citeplanguageresource{kocmi-et-al-2020-czeng}
 and 58M originally Ukrainian sentences from
 WMT NewsCrawl,\punctfootnote{\url{https://data.statmt.org/news-crawl/}}
 the Leipzig Corpora \citeplanguageresource{biemann2007leipzig},
 UberText corpus \citeplanguageresource{khaburska2019toward}
 and Legal Ukrainian Crawling by ELRC \citeplanguageresource{LegalUkrainianCrawling}.

\paragraph{Test data.}
To evaluate the system performance in the area for which it was primarily designed, i.e., the daily communication of the refugees with Czech individuals and authorities,
 we created
 two test sets:
 2,812 sentences for \ukcs{} (from March and April 2022),
 and 2,017 sentences for \csuk{} (from 2023).
We provided these two test sets to the organizers of WMT\footnote{
  Conference on Machine Translation, \url{https://www.statmt.org/wmt23/}}
 and they were published as WMT22 \ukcs{} \citep{kocmi-etal-2022-findings} and WMT23 \csuk{} \citep{kocmi-EtAl:2023:WMT}, respectively.

Some of the sentences were news crawls provided by the WMT organizers. However, most sentences were selected from the \CT{} system logs,\punctfootnote{From users who agreed to have their data used for further system development, cf. Section~\ref{sec:ethical}.} anonymized/pseudonymized and translated by professional translators.

The test sets were also annotated for (i) user type: formal (bureaucracy), news, and other (mostly individual users); and (ii) topic: general personal conversation, work, housing, transportation/travel, school and education, health, and politics.
The test sets were designed to be balanced in these respects, but also in sentence length and the number of ``noisy'' sentences,
 i.e., user-generated sentences with authentic typos, grammatical and typographic errors, and disfluencies.
The 2017 sentences (segments) of the WMT23 \csuk{} test set are separated into 5 domains (see Table~\ref{tab:csuk-domains} for evaluation):
\begin{itemize}[noitemsep,topsep=0pt,parsep=0pt,partopsep=0pt]
\item News: 567 segments from WMT news crawl,
\item Voice: 533 segments of originally spoken Czech,
 as recognized by an automatic speech recognition system (ASR),
 i.e. including some ASR errors,
\item Personal: 390 segments of personal conversation,
\item Official: 347 segments of formal/public announcements,
\item Games: 180 segments of web stories about computer games.
\end{itemize}

\paragraph{Engine.}
The backend service is run by the LINDAT/CLARIAH-CZ infrastructure.\punctfootnote{\url{https://lindat.cz}}
It uses one server with 15 CPU cores and 44 GB RAM, with the translation models for UK-CS and CS-UK loaded on 3 GPU cards: Quadro RTX 5000 + 2x GeForce RTX 2080 Ti.
The service is accessible via REST API.

The infrastructure was already set up and running for other translation pairs
when the
Russian invasion
started, so we could quickly prepare data, train new models, and provide a robust service.

\subsection{Interfaces}

We developed multiple interfaces that communicate with the translation service.

\paragraph{Original web interface.}
We have been providing the translation service for several languages even before we developed the Czech-Ukrainian models, accessible via the original web interface.%
\punctfootnote{\lindaturl{}}
Although this web interface is still available, it is more research-oriented.
For example, it enables translation via pivoting for pairs without a trained model, regardless of the final translation quality.

\begin{figure*}[!t]
    \centering
    \begin{minipage}[b]{.725\textwidth}
        \includegraphics[width=\textwidth]{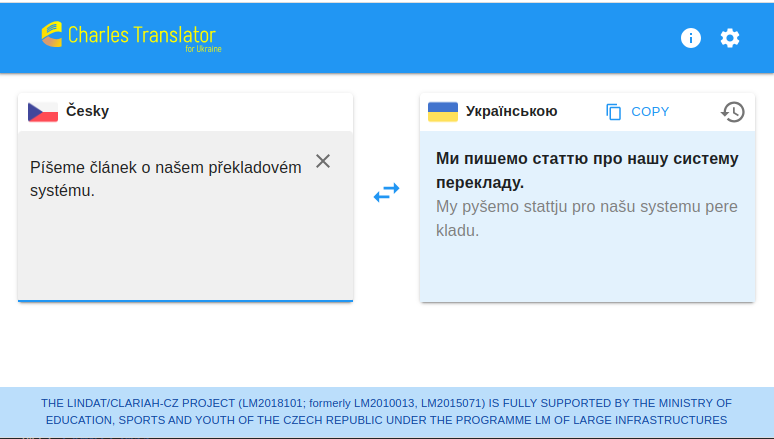}
    \end{minipage}
    \hfill
    \begin{minipage}[b]{.265\textwidth}
        \includegraphics[width=\textwidth]{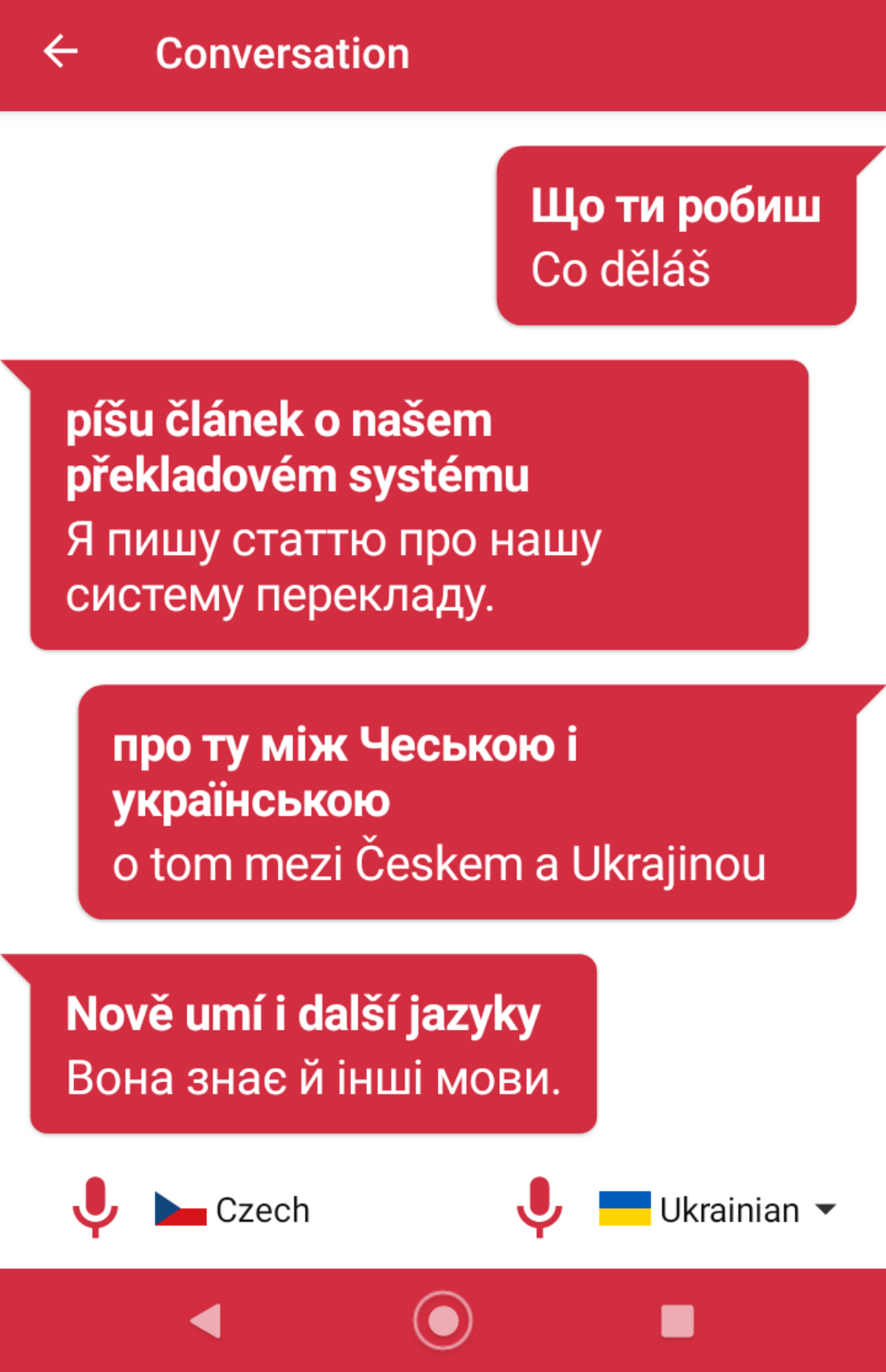}
    \end{minipage}
    \caption{Screenshots of the new web app (left) and the conversation mode in the Android app (right). Note that the Czech translation \emph{mezi Českem a Ukrajinou} (between Czechia and Ukraine) is incorrect. It should be \emph{mezi češtinou a ukrajinštinou} (between Czech and Ukrainian [language]), instead. The error is actually caused by speech recognition that capitalizes the word \ukr{чеською}, thus changing its meaning. Also, the transliteration \emph{stattju} is wrong - it should be \emph{staťťu}.
    }
    \label{fig:screenshots}
\end{figure*}



\paragraph{New web app.}
The start of the refugee crisis required a simpler application that would be accessible to end users.
To meet this demand, we developed a new React/Node.js web app that only included Ukrainian-Czech models at the time.
Following its creation on an ad-hoc organized hackathon, the development and maintenance of the app was taken over by a single developer.
Currently, the app still supports the CS$\leftrightarrow$UK translation pair only, but we plan to extend it with other pairs soon.

Because Czech and Ukrainian use different scripts, the app supports transliteration in both directions, as illustrated on the left side of Figure~\ref{fig:screenshots}.
Unlike other online translators, we use a transliteration of Ukrainian into the Latin script that is suitable for Czech speakers,
 e.g. \ukr{нашу} is transcribed as \textit{našu} instead of English-oriented \textit{nashu}.
When transcribing Czech into Cyrillic script, we keep the acute diacritic marks signaling vowel length,
 e.g. \textit{článek} is transcribed as \ukr{члáнек}.

\paragraph{Android app.}
After the web frontend was released and publicized, we extended the translation service with an Android app. 
Besides accessibility-related benefits, an Android app can take advantage
of the native speech API, offering both dictation and speech synthesis capabilities in Czech and Ukrainian.
Using these features, we also provide the app with a conversation mode, as illustrated on the right side of Figure~\ref{fig:screenshots}.
Recently, the Android app has been extended to include also other translation pairs.
The app is implemented in the Kotlin programming language using Jetpack Compose. The app is currently installed by around 2,000 people, of which 1,200 have Czech as their phone language and 600 have Ukrainian. 


\section{Quality of Translations}
\label{sec:quality}

\def\UN#1{\rowcolor{ashgrey} #1}
\def\CO#1{#1}
\begin{table}\centering
\small
\begin{tabular}{rrrr}\hline
System               & BLEU    & chrF & COMET\\\hline
\UN{GPT4-5shot}      &    32.8 & 61.0 & 90.8\\
\bf\CO{\CT{}}        &    30.2 & 57.4 & 88.0\\
\UN{GTCOM\_Peter}    &    29.8 & 57.6 & 88.9\\
\CO{CUNI-GA}         &    29.5 & 57.9 & 90.9\\
\CO{MUNI-NLP}        &    28.3 & 57.0 & 87.0\\
\UN{Lan-BridgeMT}    &    27.5 & 55.7 & 86.0\\
\UN{ONLINE-W}        &    26.8 & 55.0 & 89.4\\
\UN{ONLINE-B}        &    25.7 & 54.7 & 88.8\\
\UN{ONLINE-A}        &    25.4 & 54.4 & 88.2\\
\UN{NLLB\_MBR\_BLEU} &    25.1 & 52.3 & 86.3\\
\UN{NLLB\_Greedy}    &    24.9 & 52.5 & 86.3\\
\UN{ONLINE-G}        &    24.8 & 53.7 & 87.7\\
\UN{ONLINE-Y}        &    24.2 & 53.4 & 86.5\\
\hline
\end{tabular}
\caption{Results of automatic evaluation on the \csuk{} WMT23 test set.
Constrained systems (i.e. systems that use only the data provided by the WMT organizers) are marked with a white background.}
\label{tab:csuk}
\end{table}

\begin{table}\centering
\small
\begin{tabular}{rrrr}\hline
System              & BLEU & chrF & COMET\\\hline
\CO{AMU}            & 37.0 & 60.7 & 104.8\\
\UN{Lan-Bridge}     & 36.5 & 60.4 &  94.5\\
\UN{Online-B}       & 36.4 & 60.3 &  96.5\\
\CO{HuaweiTSC}      & 36.0 & 59.6 &  91.4\\
\UN{\CT{}-un}       & 35.9 & 59.0 &  90.2\\
\bf\CO{\CT{}}       & 35.8 & 59.0 &  88.5\\
\CO{CUNI-JL-JH}     & 35.1 & 58.7 &  89.0\\
\UN{Online-A}       & 33.3 & 57.5 &  85.4\\
\UN{Online-G}       & 31.5 & 56.3 &  84.2\\
\UN{GTCOM}          & 31.3 & 55.8 &  80.2\\
\UN{Online-Y}       & 29.6 & 55.3 &  78.6\\
\CO{ALMAnaCH-Inria} & 25.3 & 50.7 &  62.4\\
\hline
\end{tabular}
\caption{Results of automatic evaluation on the \ukcs{} WMT22 test set.
\emph{\CT{}-un} is an unconstrained version of our translator,
i.e. trained on additional training data from InterCorp. 
}
\label{tab:ukcs}
\end{table}


\begin{table*}\centering
\small
\begin{tabular}{r cc@{~}c@{~~}c@{~}c@{~}c cc@{~}c@{~~}c@{~}c@{~}c}\toprule
                     & \multicolumn{5}{c}{BLEU for domain} & \multicolumn{5}{c}{chrF for domain} \\\cmidrule(lr){2-7}\cmidrule(lr){8-13}
System               & ALL  & games & news & official & personal & voice & ALL  & games & news & official & personal & voice \\\midrule
\UN{GPT4-5shot}      & 32.8 & 26.7  & 31.8 & 36.8     & 34.5     & 32.9  & 61.0 & 57.4  & 61.9 & 63.9     & 60.1     & 59.4  \\
\bf\CO{\CT{}}          & 30.2 & 24.3  & 30.4 & 34.2     & 30.8     & 29.5  & 57.4 & 55.1  & 59.0 & 60.6     & 55.1     & 54.5  \\
\UN{GTCOM\_Peter}    & 29.8 & 24.8  & 30.7 & 35.2     & 29.5     & 26.0  & 57.6 & 54.7  & 59.6 & 61.7     & 55.1     & 53.2  \\
\CO{CUNI-GA}         & 29.5 & 24.3  & 30.6 & 33.4     & 29.7     & 26.9  & 57.9 & 55.8  & 60.0 & 60.6     & 55.8     & 54.1  \\
\CO{MUNI-NLP}        & 28.3 & 24.9  & 27.9 & 31.9     & 29.2     & 27.4  & 57.0 & 55.7  & 58.2 & 59.8     & 54.8     & 54.1  \\
\UN{Lan-BridgeMT}    & 27.5 & 24.0  & 26.9 & 31.4     & 27.8     & 26.3  & 55.7 & 54.1  & 57.6 & 58.6     & 52.7     & 52.3  \\
\UN{ONLINE-W}        & 26.8 & 20.9  & 27.3 & 32.6     & 26.1     & 24.0  & 55.0 & 51.5  & 56.9 & 59.8     & 51.8     & 51.4  \\
\UN{ONLINE-B}        & 25.7 & 20.6  & 25.0 & 31.5     & 26.4     & 24.5  & 54.7 & 52.1  & 56.2 & 58.6     & 52.2     & 51.6  \\
\UN{ONLINE-A}        & 25.4 & 20.5  & 25.1 & 30.7     & 25.8     & 23.6  & 54.4 & 51.1  & 56.0 & 58.5     & 52.3     & 51.1  \\
\UN{NLLB\_MBR\_BLEU} & 25.1 & 22.4  & 24.4 & 28.7     & 25.2     & 24.9  & 52.3 & 50.6  & 53.5 & 55.7     & 50.1     & 49.0  \\
\UN{NLLB\_Greedy}    & 24.9 & 21.7  & 25.6 & 28.1     & 24.2     & 23.5  & 52.5 & 50.8  & 54.8 & 55.6     & 49.0     & 48.6  \\
\UN{ONLINE-G}        & 24.8 & 20.6  & 25.1 & 30.9     & 24.1     & 21.1  & 53.7 & 51.3  & 55.8 & 58.3     & 50.6     & 48.5  \\
\UN{ONLINE-Y}        & 24.2 & 20.1  & 23.4 & 29.7     & 23.8     & 22.7  & 53.4 & 51.5  & 55.3 & 57.4     & 49.6     & 49.8  \\\bottomrule
\end{tabular}
\caption{BLEU and chrF results on various domains (subsets) of the \csuk{} WMT23 test set.
}
\label{tab:csuk-domains}
\end{table*}

We evaluate our system 
 on the Czech-Ukrainian WMT23\footnote{\url{https://www.statmt.org/wmt23/}} 
 General MT test set using three automatic metrics:
 BLEU~\citep{papineni-etal-2002-bleu},
 ChrF~\citep{popovic-2015-chrf}
 and COMET~\cite{rei-etal-2020-comet}.\punctfootnote{According to the human evaluation \citep{kocmi-EtAl:2023:WMT}, \CT{} (called CUNI-Transformer)
 is significantly outperformed only by systems GPT4-5shot and ONLINE-B and the human reference.}
The results in Table~\ref{tab:csuk} show different rankings for each metric, but our system is never more than 8\% worse than the best system.
Table~\ref{tab:ukcs} shows results on the Ukrainian-Czech WMT22 General MT test set.
Table~\ref{tab:csuk-domains} shows results, again on Czech-Ukrainian WMT23,
 but separately for each of the five domains included in the test set,
 including \emph{voice}, which are the ASR outputs.

While our internal manual evaluation shows notable improvements over older versions of our system,
 there are still occasional errors,
 especially in the translation of city names and personal names.\punctfootnote{%
 E.g. \textit{Košice} (a Slovak city) translates as
   \ukr{Харьков} (Russian name of the Ukrainian city \ukr{Харків}).
 }
See also the caption of Figure~\ref{fig:screenshots}
showing a translation error caused by an ASR error.

\section{Use Cases}
\label{sec:usecases}

From the very beginning of the project, we have been actively contacting organizations that we anticipated would be assisting Ukrainian refugees, asking about their translation needs.
If they were interested, we shared the \CT{} REST API with them as soon as the first version was ready.

One of the organizations that we partnered with is the Consortium of Migrants Assisting Organizations, an umbrella organization of multiple NGOs dealing with migrants.
It maintains the \emph{Pomáhej Ukrajině} (Help Ukraine) web platform,\punctfootnote{%
\url{https://www.pomahejukrajine.cz}}
which connects public offers of assistance in various areas (e.g., material aid, housing, education, leisure activities, psychological support) with the needs of individual Ukrainian refugees and organizations involved in aiding migrants.
The platform uses our API to automatically translate segments of the offers, particularly those that involve filling in free-form text.

We were contacted by the Police of the Czech Republic, who started gathering information on war crimes committed in Ukraine using an online form,\punctfootnote{\url{https://oznameni.policie.cz}} where the witnesses can report their testimonies in Czech, Ukrainian, Russian, and English.
The testimonies are highly sensitive,
 so they cannot be translated using online translators.
We thus provided the Police with the on-premise installation of our translator on their servers.

In September 2023, the translation service showed the following usage statistics:\punctfootnote{
 Excluding the usage of on-premise installations.
}
In the Ukrainian$\rightarrow$Czech translation direction, there was an average of 30,000 translation requests per day and about two million characters translated per day.
These translations were quite concise on average, with an average length of around 60 characters per request,
although there are also requests with hundreds of sentences.
In the Czech$\rightarrow$Ukrainian direction, there were approx. 12,000 requests per day and a total of approx. one million characters translated per day.

After the launch of the service in March 2022 and during its early days, the demand peaked in April 2022 at 223 million characters in the \csuk{} direction. Over the rest of 2022, it steadily decreased, with the trend continuing in 2023 from 62 million characters in January to 27 million in September.

In the \ukcs{} direction, there is no such decrease in usage. Although the highest number is 84 million characters translated in May 2022, the average use until September 2023 has been practically flat, around 65 million characters/month without any trends.

Although both Ukrainian refugees and Czech people communicating with them may need both translation directions, we hypothesize that most of the \ukcs{} translations are by Ukrainian refugees.

\section{Conclusions}
\label{sec:conclusions}

We presented \CT{}, a machine translation system between Ukrainian and Czech based on the block back-translation method. The main motivation behind its rapid development was to facilitate communication between Czech and Ukrainian speakers during the critical period following the migration from Ukraine.
The translator is available as API, web app and Android app.
\CT{}'s license allows for free use including the API for non-commercial purposes, so it can be integrated in a wide range of translating activities that are free of charge to the user.

In addition to further work on improving translation quality, we plan to adapt the model for the educational domain in order to create multilingual digital learning materials in the near future. At present, approximately 55,000 Ukrainian children are enrolled in the Czech school system, requiring a high-quality education that seamlessly integrates Ukrainian and Czech in all subjects.




\section{Acknowledgements}
This work was supported by the project TQ01000458 (EdUKate) financed by the Technology Agency of the Czech Republic (www.tacr.cz) within the Sigma 3 Programme. It was further supported by the Ministry of Education, Youth and Sports of the Czech Republic, Project No. LM2023062 LINDAT/CLARIAH-CZ and Cooperatio AMES of the Faculty of Arts, Charles University.

The authors also express their gratitude to the Institute of the Czech National Corpus (korpus.cz), to České překlady (ceskepreklady.cz) and to the following individuals for their contributions: Zdeněk Kasner, Jiří Mayer, Rudolf Rosa, Ondřej Košarko, Martin Majliš, Oxana Čmelíková, Peter Polák, Tomáš Musil, David Nápravník and Denys Bojko.



\section{Ethical considerations}\label{sec:ethical}
The sentences in the test sets (described in Section~\ref{sec:testsets}) were collected with users' opt-in consent and any personal data (except for the names of well-known public figures) were pseudonymized (using random first names and surnames), including all potentially identifying information (addresses, URLs, institution names, phone numbers, names of cities and villages). Sentences where such pseudonymization would not guarantee reasonable anonymity of the users (e.g. describing events uniquely identifying the persons involved)
 were not included in the test set.

The raw data collected are stored securely on our servers and they are not publicly available.
The users can withdraw their consent anytime in the web/mobile app and we immediately stop collecting their data (i.e. texts to be translated).
The data already stored in our database can be deleted on request.


\section{Bibliographical References}\label{sec:reference}

\bibliography{anthology-used,custom}
\bibliographystyle{lrec-coling2024-natbib}

\section{Language Resource References}
\label{lr:ref}
\bibliographystylelanguageresource{lrec-coling2024-natbib}
\bibliographylanguageresource{languageresource,anthology-used}

\end{document}